# Few-Shot Left Atrial Wall Segmentation in 3D LGE-MRI via Meta-Learning

Yusri Al-Sanaani[1], Rebecca Thornhill[1,2], Pablo Nery[3], Elena Peña[2], Robert deKemp[1,3], Calum Redpath[3], David Birnie[3], Sreeraman Rajan[1]
[1]*Department of Systems and Computer Engineering*, *Carleton University*, Ottawa, Canada
[2]*Department of Radiology, Radiation Oncology, and Medical Physics*, University of Ottawa, Ottawa, Canada
[3]*Division of Cardiology, Department of Medicine*, University of Ottawa Heart Institute, Ottawa, Canada

***Abstract***

Segmenting the left atrial (LA) wall from late gadolinium enhancement magnetic resonance imaging (LGE-MRI) is challenging because of its thin geometry, low contrast, and limited expert annotations. We propose a model-agnostic meta-learning (MAML) framework with a 3D residual U-Net backbone for K-shot (K = 5, 10, 20) LA wall segmentation. The framework is meta-trained on LA wall tasks together with auxiliary LA and right atrial (RA) cavity tasks and uses a boundary-aware composite loss to improve thin-structure delineation. We evaluated MAML on a held-out clean test set and assessed its robustness under an unseen synthetic domain shift and on a local cohort. On the held-out clean test set, MAML outperformed the K-shot fine-tuning baseline at 5-shot, achieving Dice coefficient (DSC) = 0.54 ± 0.05 versus 0.48 ± 0.06 and Hausdorff distance (HD95) = 4.60 ± 1.05 mm versus 6.40 ± 1.45 mm. At 20-shot, MAML approached the fully supervised model trained from scratch, with DSC = 0.59 ± 0.03 versus 0.61 ± 0.05. Under unseen shifts, performance decreased relative to clean testing but improved consistently as K increased. At 5-shot, MAML achieved DSC = 0.52 ± 0.06 and HD95 = 5.02 ± 1.25 mm under the unseen synthetic shift, and DSC = 0.50 ± 0.07 and HD95 = 5.43 ± 1.31 mm on the local cohort. These results suggest that meta-learning can improve thin-wall delineation in low-shot adaptation and may reduce the annotation burden for atrial remodeling assessment.

## I. INTRODUCTION

Segmenting the left atrial (LA) wall from late gadolinium enhancement magnetic resonance imaging (LGE-MRI) scans can support characterization of atrial remodeling and ablation planning in patients with atrial fibrillation [1], [2]. However, accurate LA wall segmentation remains challenging because of the wall's thin geometry, low contrast relative to adjacent tissues, and partial-volume effects, which can increase annotation variability and model uncertainty [3]–[6]. In addition, manual delineation is labor-intensive and sensitive to variations in scan quality, making high-fidelity LA wall labeling difficult to scale across institutions [6].

While deep learning has advanced cavity segmentation [7] and joint cavity-scar segmentation frameworks [8], [9], automated extraction of the thin LA wall remains a key bottleneck. Recent multicenter benchmarks report left atrial cavity Dice scores above 0.90 with surface errors below 1 mm, whereas LA wall segmentation models typically remain in the 0.60–0.70 Dice range and show larger surface-distance errors, indicating that the thin atrial shell remains a major unresolved challenge [10]–[12].

Prior reviews highlight several key challenges in LA wall segmentation methods: thin-wall morphology, open pulmonary vein/mitral valve (PV/MV) boundaries, cross-center domain variability, and the need for large labeled datasets to approach human-level segmentation performance [3]–[6]. Relative to left ventricular LGE-MRI, publicly available labeled LA LGE-MRI datasets are scarce, and expert annotation of 3D atrial structures is labor-intensive [2]. Recent mitigation strategies include 3D conditional generative models for data augmentation and task-specific transfer learning using high-accuracy cavity models as anatomical priors [13], [14]. However, scarce LA wall annotations and acquisition variability motivate few-shot adaptation.

Few-shot learning has emerged as a compelling approach to addressing label scarcity in medical image analysis [15], [16]. By leveraging prior knowledge from related tasks, models can adapt to new target structures or imaging domains using few (K) labeled training volumes. In segmentation tasks, common few-shot approaches include metric-learning methods that estimate foreground/background prototypes from support examples [17], and optimization-based meta-learning methods that learn initializations for efficient task-specific fine-tuning [15], [16].

Tang et al. proposed a prototype refinement network that improved few-shot organ segmentation by clustering and refining feature embeddings [17]. Model-agnostic meta-learning (MAML) is a gradient-based meta-learning algorithm that learns an initialization for rapid adaptation to new tasks [18]. Alsaleh et al. [15] used MAML with a 3D U-Net to demonstrate substantial gains in 5-shot settings; however, their study focused on large, distinct CT organs (e.g., liver and spleen). Khadka et al. proposed an implicit MAML algorithm for few-shot segmentation and showed 2–4% Dice gains over standard MAML on polyp and skin lesion segmentation [16]. Gama et al. applied MAML to 2D chest X-ray segmentation with simulated sparse labels, using a mini U-Net to mitigate memory costs [19]. For LGE-MRI LA segmentation, recent work has investigated few-shot approaches by learning adaptive inference mechanisms for low-contrast LA cavity segmentation [20]. Prototype methods [17] rely on the assumption that support and query features form compact clusters in a fixed embedding space. In contrast, MAML explicitly adapts model parameters through gradient-based updates, allowing the feature representation to shift during task-specific fine-tuning. This may be advantageous for LA wall segmentation, where severe partial-volume effects blur feature clusters and boundary cues, making optimization-based adaptation more suitable than prototype averaging.

Another key challenge is the robustness of LGE-MRI segmentation to domain shift. These shifts may arise from differences in scanner vendor, image acquisition protocol, image reconstruction technique, image resolution, intensity non-uniformity, noise, and site-specific preprocessing. These factors can degrade segmentation performance when models are deployed outside the training domain [6], [21], [22]. Domain generalization remains challenging in medical imaging, as models often overfit to acquisition-specific characteristics [16]. A potential benefit of meta-learning is that training on a distribution of tasks can encourage representations that are less specific to a single data source.

Recent domain generalization studies have modeled domain shift during meta-training by creating meta-train and meta-test splits from source domains. For segmentation, this has been used across distinct source domains, sites, datasets, or synthesized pseudo-modalities. These methods complement broader cross-domain robustness strategies, including data augmentation, style transfer, and normalization-based feature alignment [21], [22].

Prior few-shot segmentation studies have mainly focused on larger structures, such as the LA cavity [20] or large organs in CT [15], rather than the thin atrial wall. Robustness to domain variability has typically not been evaluated in these settings. We address these gaps by targeting the LA wall as the primary structure for few-shot meta-learning and by incorporating controlled LGE-MRI domain variations into the task construction. This task design is related to domain randomization [23], but with the controlled variations embedded within a few-shot adaptation framework. The controlled setup allows us to isolate the effect of distributional changes in line with established domain generalization practices [22].

In this paper, we propose a second-order MAML framework for K-shot 3D LA wall segmentation in LGE-MRI. The main contributions are: (i) a multi-structure meta-training scheme that leverages LA cavity and right atrial (RA) cavity annotations to learn features transferable to the thin LA wall; (ii) a controlled domain shift protocol using simulated resolution loss, bias field, intensity variation, and noise/blur perturbations, with an unseen domain held-out for meta-testing and additional evaluation on a local cohort; and (iii) a thin-structure-focused training and evaluation strategy that combines Dice, cross-entropy, and boundary loss terms and emphasizes surface-based metrics for clinically relevant boundary accuracy.

## II. Methods

### A. Meta-learning formulation

We propose a gradient-based meta-learning framework based on MAML that learns a parameter initialization for rapid adaptation to 3D LGE-MRI LA wall segmentation using only K labeled volumes. The full pipeline is depicted in Fig. 1. During the meta-training stage (Fig. 1a), the model optimizes an initialization across related segmentation tasks. During the meta-testing stage (Fig. 1b), this meta-learned initialization is fine-tuned using K labeled support volumes from a new LA wall task. The adapted model is then evaluated on disjoint query volumes from the same task (Fig. 1c). Overall, the framework meta-trains across a family of related segmentation tasks to learn atrial representations transferable to the clinically challenging LA wall under label scarcity and domain shift.

Let $\mathcal{D} = \{\mathcal{D}_{LA}, \mathcal{D}_{RA}, \mathcal{D}_{ext}\}$ denote the data sources. Each 3D LGE-MRI volume is represented as $x \in \mathbb{R}^{H \times W \times Z}$. $\mathcal{D}_{LA}$ is the 2018 LA Segmentation Challenge (LA2018) dataset with LA wall and LA cavity annotations [7]. $\mathcal{D}_{RA}$ is the RA segmentation dataset (RAS) with RA cavity annotations only [2]. $\mathcal{D}_{ext}$ is a local cohort with LA wall annotations that is reserved exclusively for evaluation and never used in meta-training or meta-validation.

We define each task as $\tau = (s, d)$, where $s$ is the target structure (segmentation class) and $d$ is the imaging domain. In this work, $s \in \{LA_{wall}, LA_{cavity}, RA_{cavity}\}$. These targets are anatomically related but differ substantially in segmentation difficulty: the cavities are larger and typically higher-contrast, while the LA wall is a thin, low-contrast structure. The domain, $d$, specifies whether images belong to the original (clean) LGE-MRI distribution or to the synthetically shifted domain conditions, as defined below. By meta-training across both structures and domains, the model is encouraged to learn features that remain useful when annotations are scarce and the test distribution differs from the training distribution.

Each task, $\tau = (s, d)$, is a binary segmentation problem for structure $s$ in images belonging to domain $d$. We sample support and query sets at the subject-level: $S_\tau = \{(x_i^d, y_i^s)\}_{i=1}^{K_\tau}$, $Q_\tau = \{(x_j^d, y_j^s)\}_{j=1}^{M_\tau}$, where $S_\tau$ and $Q_\tau$ contain disjoint subjects. The support set is used for inner-loop adaptation, whereas the query set is used for the meta-objective or evaluation. Here, $K_\tau$ is small to simulate a K-shot scenario, and $M_\tau = 10$ denotes the number of query cases used for meta-training, meta-validation, and final evaluation episodes. The support/query split follows standard episodic training for few-shot segmentation and explicitly mirrors a within-episode train/validation scenario used to compute the meta-loss [15], [16], [18].

For LA wall tasks, we set $K_\tau = K$ with $K \in \{5, 10, 20\}$. For auxiliary cavity tasks (LA cavity , RA cavity), we also enforce a small-support regime and cap the support size as $K_\tau = \min(K,\ 10)$, even when more cavity labels are available. This prevents the meta-objective from being dominated by large, easy support sets from auxiliary tasks and keeps the meta-gradient aligned with the intended few-shot adaptation setting. During meta-training, tasks are sampled as $\tau \sim p(\tau)$ over feasible $(s, d)$ pairs, with approximately uniform sampling across structures and selected training domains, and slight oversampling of LA wall tasks to reduce bias toward the easier cavity-segmentation tasks. Each episode optimizes query-set error after adapting on the support set.

### B. MAML Optimization for Segmentation

We train the model using the MAML algorithm [18]. MAML optimizes the initialization to enable rapid adaptation to new tasks. Each meta-training iteration samples a meta-batch $\mathcal{B} = \{\tau_i\}_{i=1}^{N}$, with $\tau_i \sim p(\tau)$, where N is the number of tasks per meta-batch. We use a subset-adaptation variant of MAML. The model parameters are partitioned into an inner-loop adaptable subset $\theta$, comprising the segmentation head and last decoder block, and a representation subset $\phi$, comprising the encoder and earlier layers.

For each task $\tau_i$, we start from the current initialization $\theta$ and perform one inner-loop update on its support set. The one step inner update is:

$$\theta_i' = \theta - \alpha \nabla_\theta \, \mathcal{L}_{\tau_i}\big(S_{\tau_i}; \theta, \phi\big), \tag{1}$$

where $\mathcal{L}_{\tau_i}(\cdot)$ is the segmentation loss computed on $S_{\tau_i}$ and $\alpha$ is the inner-loop learning rate.

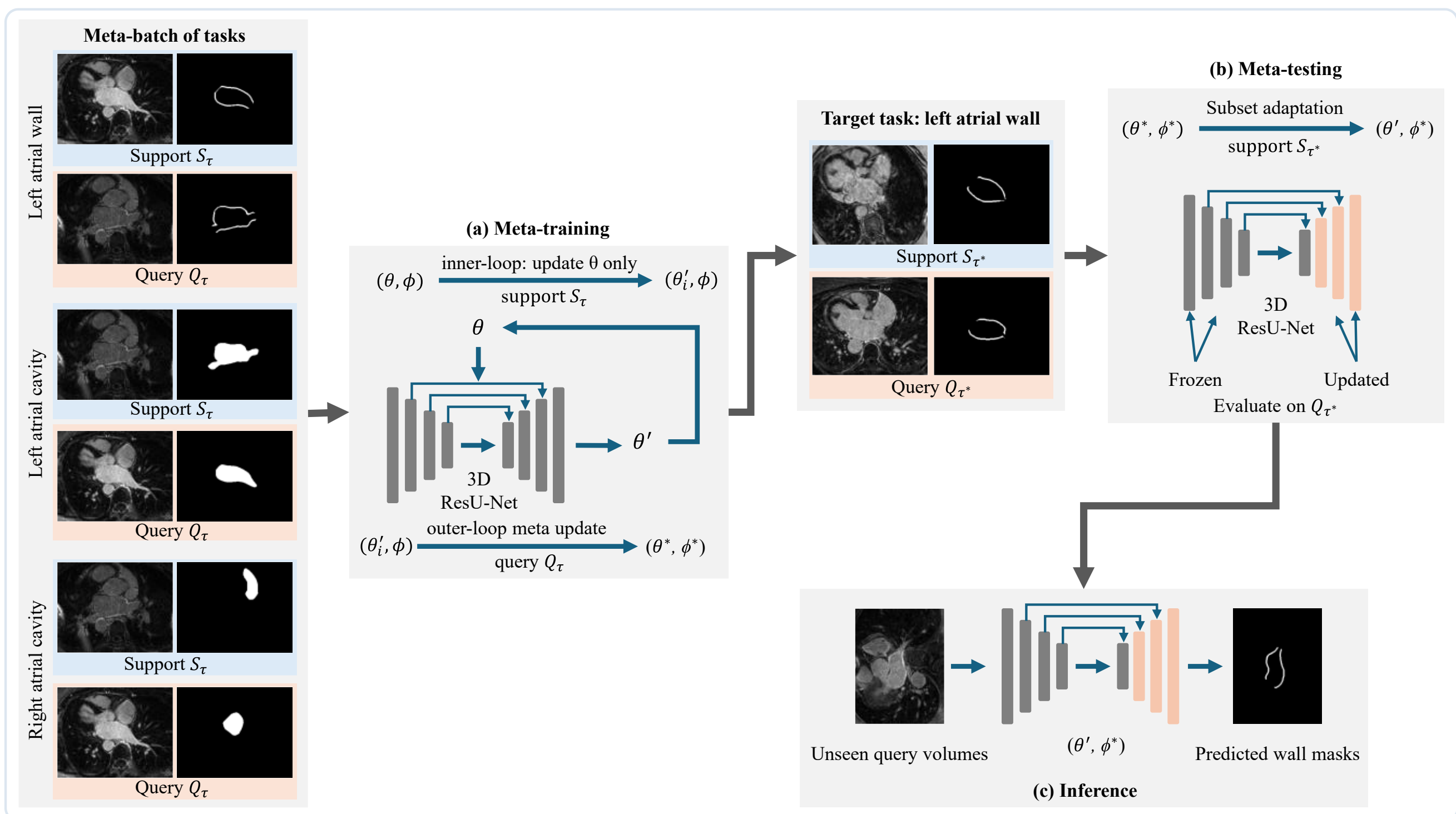

Fig. 1. Simplified overview of the proposed few-shot meta-learning framework for 3D LA wall segmentation. MAML is meta-trained on episodic LA wall and atrial cavity tasks across clean and shifted domains, then adapted with K labeled LA wall volumes and evaluated on disjoint query cases.

In our implementation, the inner-loop gradient step is applied only to the adaptable subset (head and last decoder block), while the representation parameters $\phi$ are held fixed during the inner step, as in [24]. These representation parameters are still updated in the outer loop through the query-set meta-gradient. The strategy helps preserve stability in low-data updates while still meta-learning the representation parameters over many episodes. After this task-specific adaptation, the model is evaluated on the query set $Q_{\tau_i}$. The outer loop then updates the initialization $\theta$ and the representation parameters$\phi$ to improve post-adaptation performance across all tasks in the batch:

$$\mathcal{L}_{\text{meta}}(\theta, \phi) = \sum_{i=1}^{N} \mathcal{L}_{\tau_i}\left(Q_{\tau_i}; \theta_i', \phi\right), \tag{2}$$

$$(\theta, \phi) \leftarrow (\theta, \phi) - \beta \nabla_{(\theta,\phi)} \mathcal{L}_{\text{meta}}(\theta, \phi), \tag{3}$$

with meta-learning rate $\beta$.

Second-order MAML is memory-intensive for 3D segmentation because it backpropagates through the inner-loop update. Following prior 3D MAML segmentation implementations [15], [16], we make training tractable in terms of GPU memory and runtime via: (i) patch-based learning (cropped sub-volumes), (ii) small inner-step count (n=1), (iii) gradient checkpointing, and (iv) mixed precision.

At meta-test time (adaptation), we initialize the model with the meta-learned parameters $(\theta^*, \phi^*)$ and fine-tune the model on the LA wall target task in the target domain using $K$ support volumes. This produces adapted parameters $\theta'_{\text{LA wall}}$, where $\phi$ remains fixed as $\phi = \phi^*$ under our subset-adaptation setting. We then apply the adapted model to segment query/test images from the same target domain. This design aims to provide a more effective initialization than generic initialization or unrelated pre-training for LA wall delineation under limited labels and distribution shift. The inner and test-time adaptation learning rates are selected by episodic meta-validation. We prioritize second-order MAML over first-order approximations such as Reptile [25] to maximize adaptation fidelity, given the importance of precise boundary optimization in our application, at the cost of higher computation [18].

### C. Network Architecture

We adopt a 3D residual U-Net segmentation model, inspired by nnU-Net design principles for biomedical image segmentation [26]–[28], and adapted to our meta-learning setting. The network extracts features using $f_\phi(x)$ and produces a foreground probability map through a sigmoid head: $\hat{y} = \sigma\left(g_\theta\big(f_\phi(x)\big)\right), \hat{y} \in [0,1]^{H \times W \times Z}$, where $g_\theta(\cdot)$ is the output layer and $\sigma(\cdot)$ is the sigmoid. Because each task is binary (target structure vs. background), we use a single shared output head across all tasks. This avoids a multi-class formulation under partial annotation (e.g., RA-only) and keeps the model compact, while requiring the shared initialization to encode features that transfer across atrial structures after a few-shot update.

To improve stability under extremely small support sets and to reduce overfitting risk in the 3D implementation, we adapt only a subset of parameters during the inner loop. Specifically, we adapt only the segmentation head $g_\theta$ and the last decoder block of the U-Net. Earlier encoder/decoder layers (parameterized by ϕ) remain fixed during the inner update. The subset-adaptation design preserves fast task-specific calibration while keeping adaptation well-conditioned in the low-data regime. To keep second-order meta-updates feasible, we use patch-based training on cropped 3D sub-volumes of $256^3$ voxels rather than full volumes, consistent with common practice for 3D U-Net training [2]. We use instance normalization to reduce sensitivity to intensity variation across domains. We set the dropout rate to 0.1 and use a batch size of 1. We optimize model parameters using Adam optimizer, with inner-loop learning rate $\alpha = 1 \times 10^{-4}$, outer-loop meta-learning rate $\beta = 1 \times 10^{-6}$, and a weight decay of $1 \times 10^{-5}$. These hyperparameters were guided by prior literature and further tuned empirically in our experiments [15].

### D. Loss Functions

Each task-specific segmentation loss $\mathcal{L}_\tau$ combines region-based overlap terms with a boundary-aware term suited to extreme class imbalance and thin anatomy. For a predicted probability map $\hat{y}$ and ground-truth binary mask $y$, we define:

$$\mathcal{L}_\tau = \lambda_D \mathcal{L}_{Dice} + \lambda_{CE} \mathcal{L}_{CE} + \lambda_B \mathcal{L}_{boundary}, \quad (4)$$

where $\lambda_D, \lambda_{CE}, \lambda_B$ weight the Dice, cross-entropy (CE), and boundary loss components, respectively. We use soft Dice loss to mitigate class imbalance [28]:

$$\mathcal{L}_{Dice} = 1 - \frac{2\sum_i \hat{y}_i\, y_i + \epsilon}{\sum_i \hat{y}_i + \sum_i y_i + \epsilon}, \quad (5)$$

where $\hat{y}_i \in [0,1]$, $y_i \in \{0,1\}$, $\epsilon$ is a smoothing constant (e.g., $10^{-6}$), and $i = 1, \ldots, V$ is the voxel index. We include binary cross-entropy for per-voxel supervision and to improve calibration:

$$\mathcal{L}_{CE} = -\frac{1}{V}\sum_i [\, y_i \log(\hat{y}_i) + (1 - y_i)\log(1 - \hat{y}_i)]. \quad (6)$$

Unlike Dice, CE provides stable gradients even when overlap is initially poor. In practice, combining Dice and CE losses can improve convergence compared with Dice loss alone [28], [29]. To refine thin boundaries, we incorporate the boundary loss [29]. Let $SD_Y(i)$ denote the signed distance map of the ground truth boundary at voxel $i$, with negative values inside the foreground and positive values outside. The boundary loss is:

$$\mathcal{L}_{boundary} = \frac{1}{V}\sum_i \hat{y}_i \; SD_Y(i), \quad (7)$$

which encourages probability mass to concentrate on the correct side of the ground-truth boundary and improves surface alignment for thin structures. We set the final weights to $\lambda_D = 0.60$, $\lambda_{CE} = 0.30$, and $\lambda_B = 0.10$, selected by episodic meta-validation. We apply the same composite loss during few-shot adaptation at meta-test time for consistency.

### E. Datasets and Controlled Domain Shift Protocol

We use the public LA2018 dataset [7], which contains 154 LGE-MRI volumes with LA cavity and wall annotations. Let $\mathcal{D}_{LA}^{train}$, $\mathcal{D}_{LA}^{val}$, and $\mathcal{D}_{LA}^{test}$ denote the 94/30/30 subject-level split of LA2018 for meta-training, meta-validation, and held-out meta-testing/evaluation, respectively. For auxiliary RA cavity supervision, we use the public RAS dataset [2], which contains 50 LGE-MRI volumes with RA cavity annotations only. These volumes are used only for RA cavity tasks during meta-training and are split into train/validation partitions analogously. A distinct local cohort of 42 LGE-MRI scans acquired from patients with persistent atrial fibrillation, with ground-truth annotations for the LA wall and LA cavity, is held-out entirely as a genuine unseen distribution. All data acquired from this local cohort were used following approval by the local research ethics board. All volumes are cropped and resampled to a consistent orientation and voxel spacing, with 0.625 mm isotropic spacing used for internal processing. They are then intensity-normalized per volume by clipping extreme values and scaling to zero mean and unit variance.

Real-world LGE-MRI signal intensity varies across scanners and acquisition settings, causing domain shifts that can degrade segmentation performance. To evaluate and improve robustness, we introduce a controlled domain shift protocol with synthetic transformations. We define $\mathcal{D}_{shift} = \{d_0, d_1, \ldots, d_4\}$, where $d_0$ is the clean domain and $d_1$–$d_4$ are shifted domains that mimic common LGE-MRI variations. Each domain corresponds to a transformation $T_d(x)$ applied to images, while the segmentation masks remain unchanged. The clean domain $d_0$ uses standard preprocessing and intensity normalization. The resolution-shift domain $d_1$ downsamples images in-plane (e.g., by a factor of 2), and then upsamples them back to introduce blur and partial-volume effects. The bias-field domain $d_2$ multiplies images by a smooth low-frequency field to emulate intensity inhomogeneity [30]. The contrast-variation domain $d_3$ applies gamma/histogram scaling to simulate global intensity/contrast differences. Finally, the noise-and-blur domain $d_4$ introduces additive noise and mild blur to simulate degraded image quality.

For each domain $d_k \in \{d_0, \ldots, d_4\}$, we construct domain-specific variants of the LA2018 partitions by applying $T_{d_k}$ to the images while preserving the masks. To prevent leakage, we first create a single subject-level split on the underlying clean data and then generate domain-specific variants only within the corresponding partition. Thus, each subject and all of its transformed variants belong to exactly one partition.

Meta-training samples episodes from $\mathcal{D}_{LA}^{train}(d_k)$, with $d_k \in \{d_0, d_1, d_2, d_3\}$. Episodic meta-validation uses $\mathcal{D}_{LA}^{val}(d_k)$, with $d_k \in \{d_0, d_1, d_2, d_3\}$, for hyperparameter selection and early stopping. We train on $\mathcal{D}_{train} \subset \mathcal{D}_{shift}$ and hold out at least one domain as unseen, i.e., $\mathcal{D}_{test} = \mathcal{D}_{shift} \setminus \mathcal{D}_{train}$ [21], [22]. In our primary protocol, $d_4$ is excluded from meta-training and reserved for unseen synthetic-domain testing.

During meta-training, synthetic transformations are sampled on the fly within the training domains to increase appearance diversity without reducing the effective dataset size. During evaluation, domain transforms are applied deterministically for reproducible reporting. Final evaluation is performed on held-out LA2018 subjects under the clean domain $d_0$, the unseen synthetic domain $d_4$, and the independent local cohort $d_{\text{ext}}$, which is reserved for real-world distribution-shift evaluation.

### F. *Few-shot evaluation protocol*

We evaluate few-shot performance for K∈{5, 10, 20} using an episodic meta-test protocol that mimics deployment. For each K, we sample $N_{\text{epi}} = 20$ random meta-test episodes. Each episode adapts on K labeled support volumes and is evaluated by inference on M = 10 disjoint query volumes drawn from the same meta-test pool. Within each episode, meta-testing refers to K-shot adaptation on the support set, whereas inference refers to evaluating the adapted model on the disjoint query set.

Our primary target is LA wall segmentation. We evaluate three scenarios: (a) clean in-domain evaluation, where $K$-shot adaptation and query evaluation use the held-out 30 LA wall cases under $d_0$; (b) unseen synthetic domain evaluation, where support/query volumes are drawn from $\mathcal{D}_{LA}^{test}(d_4)$; and (c) external domain evaluation, where the model adapts on $K$ labeled cases from a local cohort ($d_{\text{ext}}$) and is evaluated on $M = 10$ disjoint query cases. In all scenarios, remaining cases were available for sampling in other random episodes. Performance is averaged across $N_{\text{epi}} = 20$ random episodes for each scenario.

We compare three models. First, the K-shot supervised fine-tuning (FT) model initializes a 3D U-Net from LA cavity pre-training and then fine-tunes it on the $K$ LA wall support volumes. FT uses the same $K$-shot support sets and is evaluated on the same $M = 10$ disjoint query sets as MAML, serving as a few-shot baseline without a meta-learned initialization. Second, the proposed MAML model performs episodic meta-training across structures and training domains, followed by K-shot adaptation to the LA wall task. Finally, the fully supervised model (Full-sup) trains the same 3D U-Net using all available LA wall labels in the training partition (trained from scratch) and is reported only as a supervised upper bound, since it is not comparable in annotation cost to few-shot methods. All methods share the same backbone, preprocessing, and loss definition; hyperparameters were selected using the validation set.

### G. *Metrics*

We report Dice coefficient (DSC) as the primary overlap metric. We also report the 95th percentile Hausdorff distance (HD95, mm) to capture large boundary errors and normalized surface Dice (NSD) as a surface-based measure. For NSD, we use a 1 mm tolerance, chosen to reflect the thin anatomical scale of the LA wall. Surface metrics are emphasized because thin-wall segmentation is boundary-dominated; small surface shifts can create clinically meaningful thickness distortions while producing only modest changes in overlap [31]. All metrics are computed in 3D for each case. For few-shot methods, metrics are first averaged over the $M = 10$ query cases within each episode and then reported as mean ± standard deviation across $N_{\text{epi}} = 20$ repeated meta-test episodes. For Full-sup, metrics are reported as mean ± standard deviation across held-out test cases.

## III. Results and discussion

Table I reports the K-shot performance on the clean held-out test domain $d_0$ using unseen subjects. Support and query sets are drawn from the held-out $d_0$ pool with disjoint cases per episode. MAML achieved higher mean DSC and NSD than the FT baseline across K ∈ {5, 10, 20}, with consistently lower variability than FT. At K = 5, MAML improved DSC from 0.48 (FT) to 0.54 and NSD from 0.43 to 0.47, suggesting better overlap accuracy and surface alignment. At K = 5, HD95 decreased from 6.40 mm for FT to 4.60 mm for MAML, suggesting fewer large boundary errors. Similar trends hold at K = 10 and K = 20, where MAML maintains a consistent advantage over FT in both overlap and surface-based accuracy.

These gains are most meaningful in the low-shot regime, where wall-only training is highly sensitive to initialization and boundary ambiguity. The LA wall is thin and low-contrast, so small changes in initialization can lead to large differences in the final boundary. MAML directly optimizes the initialization for rapid post-update improvement, which is aligned with the intended K-shot deployment scenario and the original motivation of gradient-based meta-learning [18]. In practice, this optimization objective is reflected in improved mean performance at small K and reduced variance across episodes as shown in Table I. The reduced variance suggests the meta-learned initialization is less sensitive to the specific cases selected for the support set.

Relative to the fully supervised reference (Full-sup; DSC = 0.61), MAML at K = 20 reached DSC = 0.59. For boundary accuracy, MAML at K = 20 reached NSD = 0.52 and HD95 = 3.45 mm, approaching the Full-sup reference of NSD = 0.54 and HD95 = 3.12 mm.

Table II summarizes MAML's $K$-shot LA wall adaptation under two unseen shifts: the held-out synthetic domain $d_4$ and the local cohort $d_{\text{ext}}$. Relative to clean $d_0$ performance (Table I), both shifts showed lower performance, with a larger effect on surface accuracy than on overlap. On $d_4$, MAML achieved DSC = 0.52, NSD = 0.47, and HD95 = 5.02 mm at K = 5. Performance improved to DSC/NSD/HD95 = 0.55/0.51/4.20 mm at K = 10 and 0.57/0.52/3.70 mm at K = 20. This indicates that additional support labels consistently reduce worst-case boundary errors under the synthetic perturbation.

Performance on the local cohort was lower overall, with DSC/NSD/HD95 = 0.50/0.45/5.43 mm at K = 5 and 0.53/0.48/4.75 mm at K = 10. At K = 20, it approached the synthetic-shift results with DSC = 0.56, NSD = 0.50, and HD95 = 3.86 mm. Notably, the largest gap between $d_4$ and $d_{\text{ext}}$ appears in HD95 at K = 10 (4.20 vs. 4.75 mm). This pattern is consistent with thin-wall delineation being especially sensitive to real scanner/protocol differences that manifest as localized boundary outliers even when mean overlap remains comparable.

Fig. 2 shows representative 5-shot LA wall segmentations for subjects from the held-out clean domain $d_0$ and the local cohort $d_{\text{ext}}$. Compared with FT, MAML produced more continuous wall boundaries with fewer fragmented predictions, particularly in low-contrast regions. This qualitative pattern is consistent with the quantitative improvements in surface alignment and HD95 reported in Tables I and II, suggesting fewer large boundary errors in the low-shot setting.

The observed behavior suggests that the meta-learned initialization may reduce degradation under unseen domain shifts. Meta-training exposes the learner to a distribution of tasks $(s, d)$ and encourages the initialization to remain useful after a small number of adaptation steps. The auxiliary cavity tasks bias the representation toward atrial anatomy and stable shape cues, while reserving wall labels for task-specific adaptation. In other words, cavity tasks provide strong anatomical context (e.g., global atrial geometry), while MAML encourages the resulting features to remain adaptable to the wall with limited labels. This design is aligned with meta-learning practice in medical segmentation [15], [16]. Across Tables I–II, MAML's improvements come with more consistent episode-level behavior, most noticeably at low $K$. On the clean domain $d_0$ (Table I), MAML showed a smaller episode-to-episode spread than FT, especially at K = 5, suggesting reduced sensitivity to which subjects are selected for the support set.

This stability is relevant in a deployment-style K-shot protocol, where performance can fluctuate with support composition and with boundary ambiguity in low-contrast regions. We do not attribute this stability to a single mechanism. Rather, it is consistent with how the method is trained and applied. MAML optimizes post-update performance across episodes, and we further regularize low-shot adaptation by restricting inner-loop updates to the head and the last decoder block.

TABLE I. K-SHOT ADAPTATION RESULTS ON THE HELD-OUT CLEAN TEST SET ($D_0$), COMPARING K-SHOT SUPERVISED FINE-TUNING BASELINE (FT), MAML, AND THE FULLY SUPERVISED MODEL (FULL-SUP). RESULTS ARE REPORTED AS MEAN ± STANDARD DEVIATION ACROSS REPEATED META-TEST EPISODES.

| **Method** | **K = 5** | | | **K = 10** | | | **K = 20** | | |
|---|---|---|---|---|---|---|---|---|---|
| | ***DSC*** ↑ | ***NSD*** ↑ | ***HD95 (mm)*** ↓ | ***DSC*** ↑ | ***NSD*** ↑ | ***HD95 (mm)*** ↓ | ***DSC*** ↑ | ***NSD*** ↑ | ***HD95 (mm)*** ↓ |
| FT | 0.48 ± 0.06 | 0.43 ± 0.07 | 6.40 ± 1.45 | 0.50 ± 0.05 | 0.46 ± 0.06 | 5.65 ± 1.20 | 0.51 ± 0.04 | 0.49 ± 0.05 | 4.71 ± 0.98 |
| MAML | 0.54 ± 0.05 | 0.47 ± 0.06 | 4.60 ± 1.05 | 0.57 ± 0.04 | 0.50 ± 0.05 | 3.85 ± 0.90 | 0.59 ± 0.03 | 0.52 ± 0.04 | 3.45 ± 0.86 |
| Full-sup[a] | 0.61 ± 0.05 | 0.54 ± 0.06 | 3.12 ± 0.79 | 0.61 ± 0.05 | 0.54 ± 0.06 | 3.12 ± 0.79 | 0.61 ± 0.05 | 0.54 ± 0.06 | 3.12 ± 0.79 |

a. Full-sup results are repeated across K because the model does not use K-shot adaptation; values are reported as mean ± standard deviation across held-out test cases.

TABLE II. K-SHOT ADAPTATION RESULTS FOR MAML ON THE UNSEEN SYNTHETIC SHIFT ($d_4$) AND LOCAL COHORT ($d_{ext}$), REPORTED AS MEAN ± STANDARD DEVIATION ACROSS REPEATED META-TEST EPISODES.

| **Domain** | **K = 5** | | | **K = 10** | | | **K = 20** | | |
|---|---|---|---|---|---|---|---|---|---|
| | ***DSC*** ↑ | ***NSD*** ↑ | ***HD95 (mm)*** ↓ | ***DSC*** ↑ | ***NSD*** ↑ | ***HD95 (mm)*** ↓ | ***DSC*** ↑ | ***NSD*** ↑ | ***HD95 (mm)*** ↓ |
| $d_4$ | 0.52 ± 0.06 | 0.47 ± 0.07 | 5.02 ± 1.25 | 0.55 ± 0.05 | 0.51 ± 0.06 | 4.20 ± 1.00 | 0.57 ± 0.04 | 0.52 ± 0.05 | 3.70 ± 0.84 |
| $d_{ext}$ | 0.50 ± 0.07 | 0.45 ± 0.08 | 5.43 ± 1.31 | 0.53 ± 0.06 | 0.48 ± 0.07 | 4.75 ± 1.15 | 0.56 ± 0.05 | 0.50 ± 0.06 | 3.86 ± 0.95 |

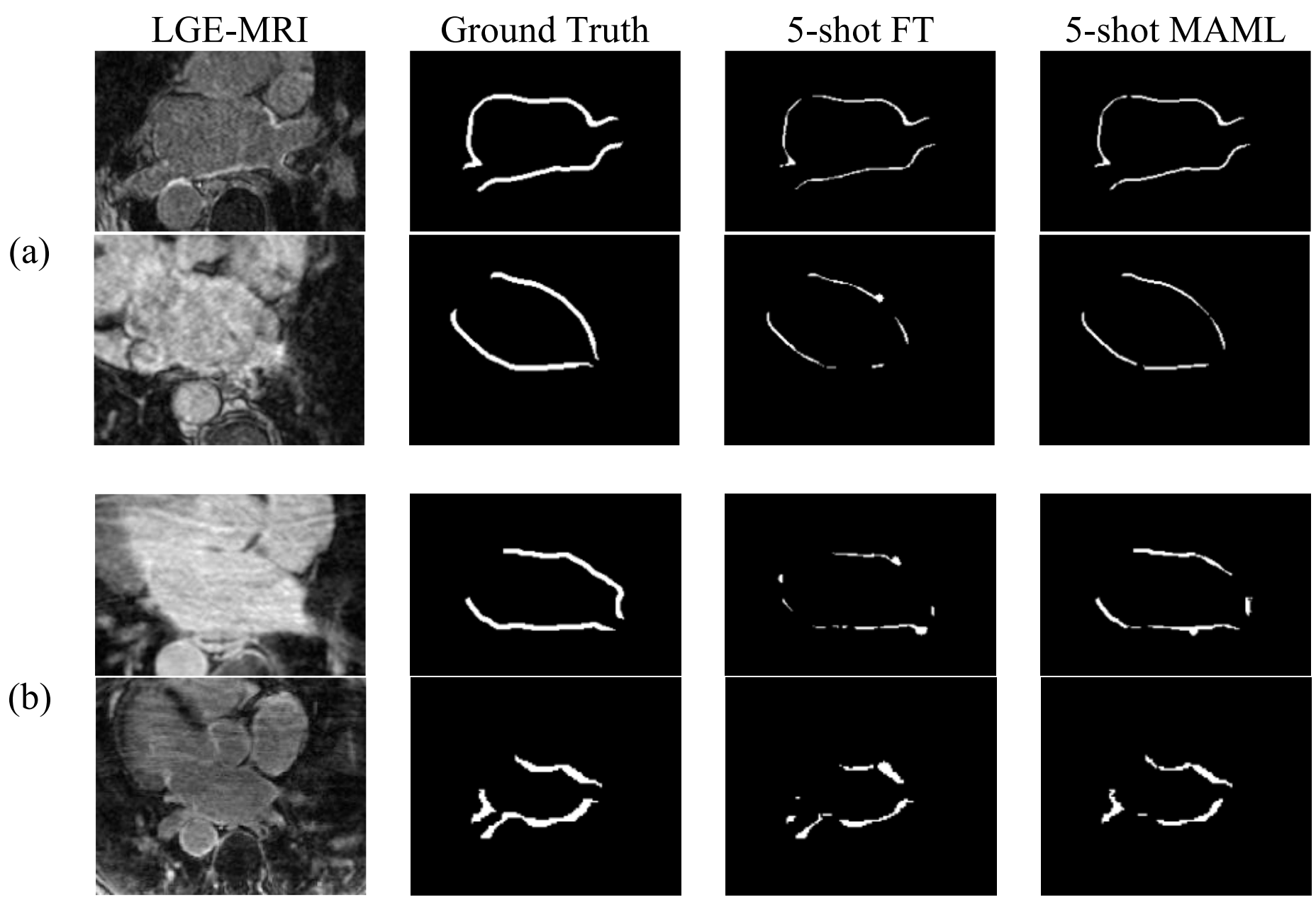

Fig. 2. Axial mid-slice comparison of LA wall segmentation with $K = 5$. Cases are drawn from (a) held-out clean domain $d_0$ subjects and (b) the local cohort $d_{\text{ext}}$.

Reduced episode-level variability is also reflected in the surface-error metrics. The tighter HD95 variability on $d_0$, together with the unseen domain shift results in Table II, suggests fewer episodes with large boundary outliers. Such stability is particularly important for thin LA wall delineation, where localized failures can dominate worst-case surface metrics even when overlap changes modestly.

The controlled domain shift protocol, while transparent and reproducible, remains a synthetic approximation of real multisite heterogeneity, including coil differences, vendor-specific reconstruction, gating variability, and population or pathology shifts. Therefore, our generalization claims are bounded to the defined shift family and the held-out evaluation partitions, with additional real-world evidence provided by the local cohort $d_{ext}$. Our design choices reflect practical constraints of 3D few-shot meta-learning. Subset adaptation (head + last decoder block) improves stability and reduces low-shot overfitting, but it can limit adaptation capacity compared with full-network fine-tuning when more labels are available, e.g., $K = 20$. Full second-order MAML improves meta-optimization fidelity but remains computationally demanding despite checkpointing and mixed precision [13], [14], [16].

Future work could therefore prioritize broader external validation across multiple institutions and scanners. It could also adopt more rigorous robustness assessments (e.g., leave-one-domain-out across $d_1 \ldots d_4$) to quantify sensitivity to unseen shifts under a standardized protocol. Finally, LA wall delineation is boundary-dominated and limited by partial-volume effects. Future work could therefore investigate thin-structure-tailored objectives, such as signed-distance representations, thickness-consistency regularization, and surface-based losses that better align training with NSD and HD95.

## IV. Conclusion

Quantifying LA wall remodeling is clinically important for characterizing atrial cardiomyopathy and informing ablation planning in atrial fibrillation. However, accurate wall delineation in LGE-MRI remains difficult because of the wall's thinness, low contrast, and sensitivity to acquisition variability. In this work, we presented a second-order MAML framework for label-efficient 3D LA wall segmentation. The framework uses related atrial structures, including the LA wall and the LA and RA cavities, together with controlled training domain variations to learn an initialization that supports rapid K-shot adaptation. Across clean domain and unseen domain shift evaluations, including a held-out synthetic shift and a distinct local cohort, the meta-learned initialization improved overlap and surface accuracy compared with the FT baseline in the low-shot regime. It also showed more reliable episode-level behavior, reflected by lower HD95 values and reduced variability. These findings suggest that meta-learning may reduce dependence on large, center-specific LA wall datasets and support practical adaptation to new clinical environments using only a small number of annotated cases. This may lower the annotation burden required to adapt automated atrial analysis methods for multicenter studies and future clinical workflows.

## Acknowledgment

This research was enabled in part by the Natural Sciences and Engineering Research Council of Canada (NSERC) Discovery Grant, and by computational resources provided by the Digital Research Alliance of Canada (https://alliancecan.ca).

## References

[1] Y. Wu, Z. Tang, B. Li, D. Firmin, and G. Yang, "Recent advances in fibrosis and scar segmentation from cardiac MRI: A state-of-the-art review and future perspectives," *Front. Physiol.*, vol. 12, Art. no. 709230, Aug. 2021, doi: 10.3389/fphys.2021.709230.

[2] J. Zhu et al., "RAS dataset: A 3D cardiac LGE-MRI dataset for segmentation of right atrial cavity," *Sci. Data*, vol. 11, Art. no. 401, Apr. 2024, doi: 10.1038/s41597-024-03253-9.

[3] A. S. Ismail, Y. Baghdady, M. A. Salem, and A. A. Wahab, "The use of MRI in quantification of the atrial fibrosis in patients with rheumatic mitral disease," *Egypt. J. Radiol. Nucl. Med.*, vol. 51, Art. no. 199, Sep. 2020, doi: 10.1186/s43055-020-00322-y.

[4] J. Ma, Q. Chen, and S. Ma, "Left atrial fibrosis in atrial fibrillation: Mechanisms, clinical evaluation and management," *J. Cell. Mol. Med.*, vol. 25, no. 6, pp. 2764–2775, Mar. 2021, doi: 10.1111/jcmm.16350.

[5] L. Li, V. A. Zimmer, J. A. Schnabel, and X. Zhuang, "Medical image analysis on left atrial LGE MRI for atrial fibrillation studies: A review," *Med. Image Anal.*, vol. 77, Art. no. 102360, Apr. 2022, doi: 10.1016/j.media.2022.102360.

[6] H. Lin et al., "Usformer: A small network for left atrium segmentation of 3D LGE MRI," *Heliyon*, vol. 10, no. 7, Art. no. e28539, Apr. 2024, doi: 10.1016/j.heliyon.2024.e28539.

[7] Z. Xiong et al., "A global benchmark of algorithms for segmenting the left atrium from late gadolinium-enhanced cardiac magnetic resonance imaging," *Med. Image Anal.*, vol. 67, Art. no. 101832, Jan. 2021, doi: 10.1016/j.media.2020.101832.

[8] G. Yang et al., "Simultaneous left atrium anatomy and scar segmentations via deep learning in multiview information with attention," *Future Gener. Comput. Syst.*, vol. 107, pp. 215–228, Jun. 2020, doi: 10.1016/j.future.2020.02.005.

[9] L. Li, V. A. Zimmer, J. A. Schnabel, and X. Zhuang, "AtrialJSQnet: A new framework for joint segmentation and quantification of left atrium and scars incorporating spatial and shape information," *Med. Image Anal.*, vol. 76, Art. no. 102303, Feb. 2022, doi: 10.1016/j.media.2021.102303.

[10] C. Thesing, A. Bueno-Orovio, and A. Banerjee, "Evaluating convolution, attention, and Mamba based U-Net models for multi-class bi-atrial segmentation from LGE-MRI," in *Lect. Notes Comput. Sci.*, vol. 15448, 2025, pp. 214–225, doi: 10.1007/978-3-031-87756-8_22.

[11] E. Almar-Munoz et al., "Multi-loss 3D segmentation for enhanced bi-atrial segmentation," in *Lect. Notes Comput. Sci.*, vol. 15448, 2025, pp. 236–244, doi: 10.1007/978-3-031-87756-8_24.

[12] A. Zolotarev, K. Johnson, A. Khan, G. Slabaugh, and C. Roney, "An ensemble of 3D residual encoder UNet models for solving multi-class bi-atrial segmentation challenge," in *Lect. Notes Comput. Sci.*, vol. 15448, 2025, pp. 209–213, doi: 10.1007/978-3-031-87756-8_21.

[13] Y. Al-Sanaani, R. Thornhill, and S. Rajan, "3D conditional image synthesis of left atrial LGE MRI from composite semantic masks," in *Proc. IEEE Int. Conf. Imaging Syst. Techn. (IST)*, 2025, pp. 1–6, doi: 10.1109/IST66504.2025.11268401.

[14] Y. Al-Sanaani, R. Thornhill, and S. Rajan, "C2W-Tune: Cavity-to-wall transfer learning for thin atrial wall segmentation in 3D LGE-MRI," in *Proc. Int. Conf. Artif. Intell. Med.*, in press.

[15] A. M. Alsaleh, E. Albalawi, A. Algosaibi, S. S. Albakheet, and S. B. Khan, "Few-shot learning for medical image segmentation using 3D U-Net and model-agnostic meta-learning (MAML)," *Diagnostics*, vol. 14, no. 12, Art. no. 1213, Jun. 2024, doi: 10.3390/diagnostics14121213.

[16] R. Khadka et al., "Meta-learning with implicit gradients in a few-shot setting for medical image segmentation," *Comput. Biol. Med.*, vol. 143, Art. no. 105227, Apr. 2022, doi: 10.1016/j.compbiomed.2022.105227.

[17] S. Tang et al., “Few-shot medical image segmentation with high-fidelity prototypes,” *Med. Image Anal.*, vol. 100, Art. no. 103412, Feb. 2025, doi: 10.1016/j.media.2024.103412.

[18] C. Finn, P. Abbeel, and S. Levine, “Model-agnostic meta-learning for fast adaptation of deep networks,” in *Proc. 34th Int. Conf. Mach. Learn. (ICML)*, vol. 70, 2017, pp. 1126–1135.

[19] P. H. T. Gama, H. Oliveira, and J. A. dos Santos, “Learning to segment medical images from few-shot sparse labels,” in *Proc. 34th SIBGRAPI Conf. Graphics, Patterns and Images*, 2021, pp. 89–96, doi: 10.1109/SIBGRAPI54419.2021.00021.

[20] J. Chen et al., “Adaptive dynamic inference for few-shot left atrium segmentation,” *Med. Image Anal.*, vol. 98, Art. no. 103321, Dec. 2024, doi: 10.1016/j.media.2024.103321.

[21] H. Guan and M. Liu, “Domain adaptation for medical image analysis: A survey,” *IEEE Trans. Biomed. Eng.*, vol. 69, no. 3, pp. 1173–1185, Mar. 2022, doi: 10.1109/TBME.2021.3117407.

[22] J. S. Yoon, K. Oh, Y. Shin, M. A. Mazurowski, and H.-I. Suk, “Domain generalization for medical image analysis: A review,” *Proc. IEEE*, vol. 112, no. 10, pp. 1583–1609, Oct. 2024, doi: 10.1109/JPROC.2024.3507831.

[23] Z. Ye, K. Wang, W. Lv, Q. Feng, and L. Lu, “FSDA-DG: Improving cross-domain generalizability of medical image segmentation with few source domain annotations,” *Med. Image Anal.*, vol. 105, Art. no. 103704, Oct. 2025, doi: 10.1016/j.media.2025.103704.

[24] A. Raghu, M. Raghu, S. Bengio, and O. Vinyals, “Rapid learning or feature reuse? Towards understanding the effectiveness of MAML,” in *Proc. Int. Conf. Learn. Represent. (ICLR)*, 2020.

[25] A. Nichol, J. Achiam, and J. Schulman, “On first-order meta-learning algorithms,” 2018, *arXiv:1803.02999*. [Online]. Available: https://arxiv.org/abs/1803.02999

[26] O. Ronneberger, P. Fischer, and T. Brox, “U-Net: Convolutional networks for biomedical image segmentation,” in *Lect. Notes Comput. Sci.*, vol. 9351, 2015, pp. 234–241, doi: 10.1007/978-3-319-24574-4_28.

[27] Ö. Çiçek, A. Abdulkadir, S. S. Lienkamp, T. Brox, and O. Ronneberger, “3D U-Net: Learning dense volumetric segmentation from sparse annotation,” in *Lect. Notes Comput. Sci.*, vol. 9901, 2016, pp. 424–432, doi: 10.1007/978-3-319-46723-8_49.

[28] F. Isensee, P. F. Jaeger, S. A. A. Kohl, J. Petersen, and K. H. Maier-Hein, “nnU-Net: A self-configuring method for deep learning-based biomedical image segmentation,” *Nat. Methods*, vol. 18, no. 2, pp. 203–211, Feb. 2021, doi: 10.1038/s41592-020-01008-z.

[29] H. Kervadec et al., “Boundary loss for highly unbalanced segmentation,” *Med. Image Anal.*, vol. 67, Art. no. 101851, Jan. 2021, doi: 10.1016/j.media.2020.101851.

[30] A. T. Simkó, T. Löfstedt, A. Garpebring, T. Nyholm, and J. Jonsson, “MRI bias field correction with an implicitly trained CNN,” in *Proc. Mach. Learn. Res.*, vol. 172, 2022, pp. 1125–1138.

[31] S. Nikolov et al., “Clinically applicable segmentation of head and neck anatomy for radiotherapy: Deep learning algorithm development and validation study,” *J. Med. Internet Res.*, vol. 23, no. 7, Art. no. e26151, Jul. 2021, doi: 10.2196/26151.